\documentclass[conference]{IEEEtran}
\usepackage{bm}
\usepackage{amsmath}
\usepackage{amssymb}
\usepackage{caption}
\usepackage{multirow}
\usepackage{times}
\usepackage{epsfig}
\usepackage{placeins}
\usepackage{colortbl}
\usepackage{subcaption}
\usepackage{graphicx}
\usepackage{comment}
\usepackage{pifont}
\usepackage{color}
\usepackage{units}
\usepackage{amsthm}
\usepackage{slashbox}
\usepackage{booktabs}

\usepackage[table, svgnames, dvipsnames]{xcolor}

\usepackage{cite}

\usepackage{tikz,pgf,pgfplots}
\usetikzlibrary{patterns}
\usetikzlibrary{shapes,arrows}
\usetikzlibrary{plotmarks}
\usetikzlibrary{external} 
\usetikzlibrary{decorations.pathmorphing}
\usetikzlibrary{arrows.meta}
\usetikzlibrary{positioning,fit}

\ifCLASSINFOpdf
\else
\fi
\begin{document}
\title{Investigation of Uncertainty of Deep Learning-based Object Classification on Radar Spectra}

\author{\IEEEauthorblockN{Kanil Patel$^{1,2}$,
William Beluch$^{1}$,
Kilian Rambach$^{1}$, 
Adriana-Eliza Cozma$^{1}$,
Michael Pfeiffer$^{1}$,
Bin Yang$^{2}$
} \\
\IEEEauthorblockA{
$^{1}$Bosch Center for Artificial Intelligence, Renningen, Germany \\
$^{2}$Institute of Signal Processing and System Theory, University of Stuttgart, Stuttgart, Germany \\
}}

\newcommand{\dan}[1]{\textcolor{red}{Dan: #1}}
\newcommand{\kanil}[1]{\textcolor{blue}{Kanil: #1}}
\newcommand{\bill}[1]{\textcolor{green}{Bill: #1}}
\newcommand{\MP}[1]{\textcolor{magenta}{MP: #1}}

\maketitle
\IEEEpeerreviewmaketitle

\begin{abstract}

Deep learning (DL) has recently attracted increasing interest to improve object type classification for automotive radar.
In addition to high accuracy, it is crucial for decision making in autonomous vehicles to evaluate the reliability of the predictions; however, decisions of DL networks are non-transparent.
Current DL research has investigated how uncertainties of predictions can be quantified, and in this article, we evaluate the potential of these methods for safe, automotive radar perception.
In particular we evaluate how uncertainty quantification can support radar perception under (1) domain shift, (2) corruptions of input signals, and (3) in the presence of unknown objects.
We find that in agreement with phenomena observed in the literature, deep radar classifiers are overly confident, even in their wrong predictions. 
This raises concerns about the use of the confidence values for decision making under uncertainty, as the model fails to notify when it cannot handle an unknown situation.
Accurate confidence values would allow optimal integration of multiple information sources, e.g. via sensor fusion.
We show that by applying state-of-the-art post-hoc uncertainty calibration, the quality of confidence measures can be significantly improved, thereby partially resolving the over-confidence problem.
Our investigation shows that further research into training and calibrating DL networks is necessary and offers great potential for safe automotive object classification with radar sensors.

\end{abstract}

\section{Introduction}
Perceiving the environment through accurate object detection and classification is a prerequisite for safe and robust decision making in automated driving.
Their robustness to weather and lighting conditions makes radar sensors a crucial component of the sensor set, but existing perception algorithms have not yet reached the performance level required for fully autonomous systems.
Recently, deep learning has been suggested for classifying objects from radar signals, because of its ability to learn highly discriminative features, which are often not easily recognized by domain experts~\cite{patel_radar19}.

Deep neural networks (DNNs), however, are predominantly treated as non-transparent black-box models, and failures are hard to detect and explain.
This is particularly concerning in safety critical applications like automated driving, where reliance on the superior perception capabilities of DNNs might result in catastrophic failures when the model is unsure or wrong.
Therefore, it is crucial to also quantify the uncertainty associated with each prediction.
This allows any upstream decision making module the opportunity to combine all relevant information from the entire sensor set and additional prior knowledge in an optimal fashion.

Although DNNs typically have softmax-outputs that provide a pseudo-confidence for each prediction, it is known that they predict values that are not usable as true measures of uncertainty, and require uncertainty calibration~\cite{pmlr-v70-guo17a}.
Without uncertainty calibration, DNNs often exhibit extreme overconfidence in their predictions, regardless of their correctness, as well as for unknown inputs~\cite{hein2019relu, kristiadi2020being}.
A well-calibrated network, instead, will associate every output with an uncertainty measure that approximates the probability of misclassfication.

In this paper we investigate for the first time the weaknesses of DNNs in providing accurate confidence measures for radar object type classification, and the potential consequences.
Our work focuses on the utility of computationally efficient post-hoc uncertainty calibration methods to improve the confidence predictions. 
Such solutions can be easily applied to already trained networks, and require no further retraining of the network parameters. 
We additionally investigate the role that the construction of the test set plays in detecting over-confidence.
If training and test splits are similar in their distribution, seemingly high accuracy might be misleading in evaluating the true generalization performance on unseen real-world data.

Radar perception requires dealing with multiple factors that introduce uncertainty, e.g.
1) corrupted signals due to noise, interference, and environmental effects,
2) ambiguities between different objects when viewed from different angles and distances, 
3) variability within different instances of the same object class, and
4) domain shifts between the datasets used for training and testing.
As radar signals are hard to interpret for humans, it is crucial to have objective measures that quantify the effects of such confounding factors.

We focus our experiments on evaluating the accuracy and confidence of DNNs which classify $7$ different categories of objects.
For training and testing we recorded two automotive datasets that vary in the environmental conditions during recording, include different driving patterns of the recording vehicle, and contain different instances of the same object class (e.g. different models of cars).
This allows a thorough comparison of the generalization capabilities of DNNs under realistic domain shifts between training and test datasets.

Additional tests investigate how artificially introduced input corruptions, which attempt to mimic expected real-world perturbations, affect both accuracy and confidence estimates.
As expected, stronger corruptions reduce the accuracy of DNNs, but we also observe an increasing deterioration of confidence estimates.
Finally, tests show that the over-confidence phenomena remains even for objects never previously seen.

Observations reported in this article highlight that current DL approaches for radar classification struggle to produce trustworthy confidence estimates for their predictions, even if on average their classification accuracy is very high.
This indicates that further research is needed to provide downstream decision making modules with high quality uncertainty estimates, in addition to highly accurate predictions.
Our work suggests that accuracy on a fixed test set alone is insufficient to predict the performance under real-world conditions, and calibrated confidence measures, in particular on corrupted and unseen data, should always be evaluated in conjunction.

\section{Related Work}

\newcommand{\cf}{cf.\ }
\newcommand{\ie}{i.\,e.\ }
\newcommand{\eg}{e.\,g.\ }
\newcommand{\wrt}{w.\,r.\,t.\ }
Various concepts have been suggested for automotive radar classification.
A common approach is to first detect reflections in  radar spectra, \eg by using a constant false alarm rate (CFAR) detector \cite{rohlingCFAR}, followed by the computation of different reflection attributes such as range and azimuth.
Features derived from these reflections can be used to exploit classification~\cite{schubert2015-dbscan, Scheiner_2019, prophet2018_car_dog_ped_classifier}.
In \cite{kraus2020GhostImagesRadar,schumann2018SemanticSegmentationRadar}, deep learning is directly applied to the radar reflections. 
Alternatively, the authors of \cite{ensembleLombacher2017,staticObjectClassificationDL-lombacher2016} create occupancy grids from radar reflections, and classify them using deep neural networks.

In an attempt to leverage as much information as possible, without any loss owed to the detection stage, classification is also performed directly on the radar spectra.
Manual features based on the spectra are determined and used for classification of different objects in \cite{rohling-car-pedestrian-features-2010, bartsch2012-pedestrian-recognition,prophet-image-based-pedestrian-classification2018}.
Data-driven approaches are investigated in \cite{convLSTM-Radar-microDoppler2019}, where the range-Doppler spectrum over multiple cycles is computed.
The authors use this as input to a combination of a convolutional neural network (CNN) and Long-Short-Term-Memory (LSTM) for classification of bicyclists and pedestrians.
The work in \cite{patel_radar19} uses a CNN on regions of interest of the range azimuth-spectra to classify different kinds of stationary objects. 
Detection and classification of different moving road users using a DNN on the range-azimuth-Doppler spectra is investigated in \cite{objectDetectionRadarSpectra2020}, and \cite{3dRadarcube-Cnn-classification2020} combines radar reflections and parts of the radar spectra as input to a neural network for classification of moving objects.

Uncertainty calibration for DNNs is an active field of research in the machine learning community.
One family of approaches focuses on generating uncertainty via multiple model predictions \cite{DropoutGalG16, EnsemblesNIPS2017}.  While powerful, such methods rely on computing multiple forward passes (via sampling or ensembles) at inference time.  For automotive applications this creates unacceptable latencies, thus more computationally efficient methods are needed.
Sampling-free uncertainty estimation \cite{Postels_2019_ICCV} and data augmentation \cite{patel2019onmanifold, OnMixupTrainThul} are candidates. 
In comparison to the methods above that modify the training process, a simple approach that requires no retraining of the models, is post-hoc calibration \cite{pmlr-v70-guo17a}.
Such methods operate on the network outputs and use a new validation set to optimize various criteria directly focused on improving uncertainty calibration~\cite{pmlr-v70-guo17a, wenger2020calibration, patel2020imax}.

\section{Setup and Methodology}
\label{setup_methodology}

\subsection{Radar Sensor, Ground Truth and Data Pre-Processing}
We use the same measurement framework, setup and automotive radar sensor as described in~\cite{patel_radar19} to measure, collect, pre-process, label, and extract the regions-of-interests (ROIs) from radar spectra.
Detailed information can be found in~\cite{patel_radar19}.

\subsection{Environment Setup}
We construct two different scenes using the following set of seven objects (commonly found in urban scenarios): car, construction barrier, motorbike, baby carriage, bicycle, pedestrian, and stop sign. 
Both scenes are static and only the radar sensor mounted on the front bumper of the test vehicle moves through the scene in various driving patterns.

The two scenes differ in the relative locations of the objects and in the driving patterns through the scene.
To increase the contrast between the two scenes, we also use different instances of some objects (different models of cars, bicycles, motorbikes and pedestrians).
We split these two scenes into two different datasets, namely: Env1 and Env2.
We \emph{only} use some data from Env1 for training and leave Env2 for an unseen evaluation of the performance of the networks based on multiple metrics.
Env1 includes multiple repetitions of all the driving patterns.
We further split  Env1 into three datasets based on the repetitions: using 1 repetition each to construct the validation set (Env1-Valid) and test set (Env1-Test) and the rest for the training set (Env1-Train).
The number of samples in each dataset (capturing objects up to $30$m) are as follows:  $192186$, $2446$, and $22954$  for Env1-Train, Env1-Valid, and Env1-Test, respectively, and $33895$ samples for Env2. 

It is important to note that the Env1-* datasets differ in measurement repetitions (i.e. measured at different time instances), but still share many of the same characteristics in the data.
This is reflected in the network's ability to generalize well to the Env1-Test, as the distribution is very similar to the training data, compared to the relatively weaker performance when evaluating on the dataset Env2.
In other words, in a different scene the network is still highly accurate, but less than when the scene setting is matched during training.

Due to the scarcity of labeled data for radar networks, often times only a single dataset (from one scene) is available for learning and evaluating the classifiers (e.g. Env1).
As a consequence it becomes important to pay close attention to the train-test split. 
Having large overlap between the two splits (e.g. naively randomly splitting frames from the same driving sequence) might lead to misguided conclusions about the generalization capabilities.
For example, consecutive frames measured milliseconds apart from one another in a static scene will have almost identical spectra (with some minor stochastic differences).
Using this kind of split will over-estimate the classifier's generalization performance.
These kind of situations require special careful consideration of the split, and more importantly also require the need to evaluate measures other than pure accuracy (e.g. to detect over-fitting).

Additionally, we propose to measure the performance of the classifier on perturbations in the input as an alternative performance measure.
In addition to accuracy, quantifying the degree to which the classifier is able to increase its uncertainty given unseen corruptions in the input is also important.

In order to study the behavior of the network on outlier object types, we collect ROI spectra of objects not used during training.
These objects range from various metallic objects to sand bags and concrete blocks. 
In total we use $5$ objects to construct a special out-of-distribution (OOD) dataset.
For the OOD dataset, the ideal behavior of the networks is to assign uniform confidence across all classes or at least provide a low-confidence prediction of the most similar class, as none of these $5$ outlier objects are seen during training (i.e. ideally we want to be able to abstain from prediction in such scenarios).

\section{Uncertainty Metrics}
The quality of the confidence estimates for neural networks can be evaluated using the Expected Calibration Error (ECE)~\cite{pmlr-v70-guo17a} and Mean Maximal Confidence (MMC)~\cite{hein2019relu}.

The ECE gives an indication of the correctness of the confidence estimates output by the network.

Given a set of samples, $S$, \emph{all} assigned a confidence of $C_{S}$ by a classifier, the classification accuracy of the samples in this set, $A_{S}$,  should match the confidence (i.e. $C_{S} = A_{S}$).
If $C_{S} < A_{S}$, then the classifier is said to be under-confident, whereas if $C_{S} > A_{S}$ the classifier is said to be over-confident.  
The ECE is a weighted average of the differences between confidence and accuracy over multiple such sets $S$.
These sets are created by binning the output probability space, and is thus optimized (i.e. ECE=0) by a classifier's confidence matching its accuracy. 
In practice, this is calculated by grouping samples into discrete bins $b$. 
More formally, ECE is computed as:
\begin{equation}
\mathrm{ECE} = \sum_{r=1}^{N_{B}} \frac{N_{b}}{N} | \mathrm{accuracy}(b) - \mathrm{confidence}(b)| ,
\end{equation}
for $N$ samples, and the calibration range $b$ is defined by the $[N/B]^{\mathrm{th}}$ index of the sorted predicted confidences.

The MMC is the mean confidence across a set of samples, with the confidence being the maximum predicted class probability (i.e. confidence of the predicted class). 
More formally, for a set of samples $S$, the MMC is given as  $\frac{1}{|S|}\sum_{1}^{|S|}\max P(y|x^{*}, W)$, where $P(y|x^{*}, W)$ is the class probability vector for a given input $x^{*}$ and parameters $W$.  
We examine the MMC of correct vs. incorrect classifications (or vs. outlier samples), for which we wish the former set to have a high MMC, and the latter sets to have a low MMC.

\section{Confidence Improving Methods}
A family of approaches, to improve confidence estimates of DNNs, that are relatively simple to implement are post-hoc uncertainty calibration methods, in which the confidences output by the network are modified by some algorithm, optimized on the validation dataset.  
Importantly, these approaches leave the network parameters unchanged (i.e. no retraining).  

The simplest approach, temperature scaling (TS)~\cite{pmlr-v70-guo17a}, uniformly scales all confidences by a learned temperature $T$, obtained by optimizing the negative log-likelihood on the validation set (in effect uniformly reducing/increasing confidence). 
While extremely simple to use, this approach has limited expressive power, and more recent approaches yield better calibration gains. 
Two such methods are used in this paper to demonstrate the effectiveness of post-hoc methods to improve network calibration.  
The first is an improved scaling method using a latent Gaussian processes (GP)~\cite{wenger2020calibration}, and the second is a discrete binning method based on maximizing the mutual information between the network outputs and the labels, I-Max~\cite{patel2020imax}.
It should be noted that these families of methods take as input the predicted probabilities of the deep network and further refine them to generate more accurate confidences.

\section{Experiments}

\subsection{Network Training}
Radar ROI spectra were used to train a Convolutional Neural Network (CNN)~\cite{lecun1995convolutional}.
The CNN consists of $3$ convolutional layers, containing $16$, $32$ and $64$  $3\times3$ filters, in addition to $2\times2$ max-pooling layers.
They are followed by $2$ fully-connected layers (with $512$ and $32$ neurons), each using Batch Normalization~\cite{batchnorm} and Dropout~\cite{dropout}.

We train the networks only using Env1-Train and select the model with the best accuracy on the independent validation set Env1-Valid.
We evaluate the performance of the networks on two unseen test datasets: Env1-Test and Env2-Test.
We train $10$ independent networks with different initializations and report their mean and standard deviations.

\begin{figure} [!htp]
\centering
\begin{subfigure}{0.95\columnwidth}
  \centering
  \includegraphics[width=1.0\textwidth, clip]{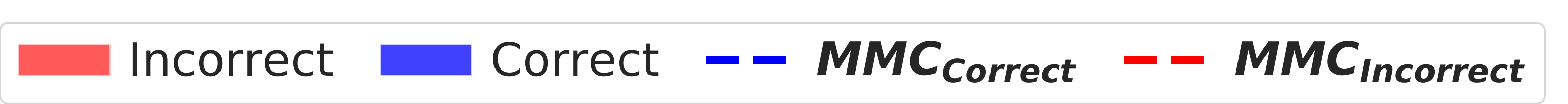}
  \label{legend}
\end{subfigure}
\begin{subfigure}{.492\columnwidth}
  \centering
  \includegraphics[width=0.95\textwidth, clip]{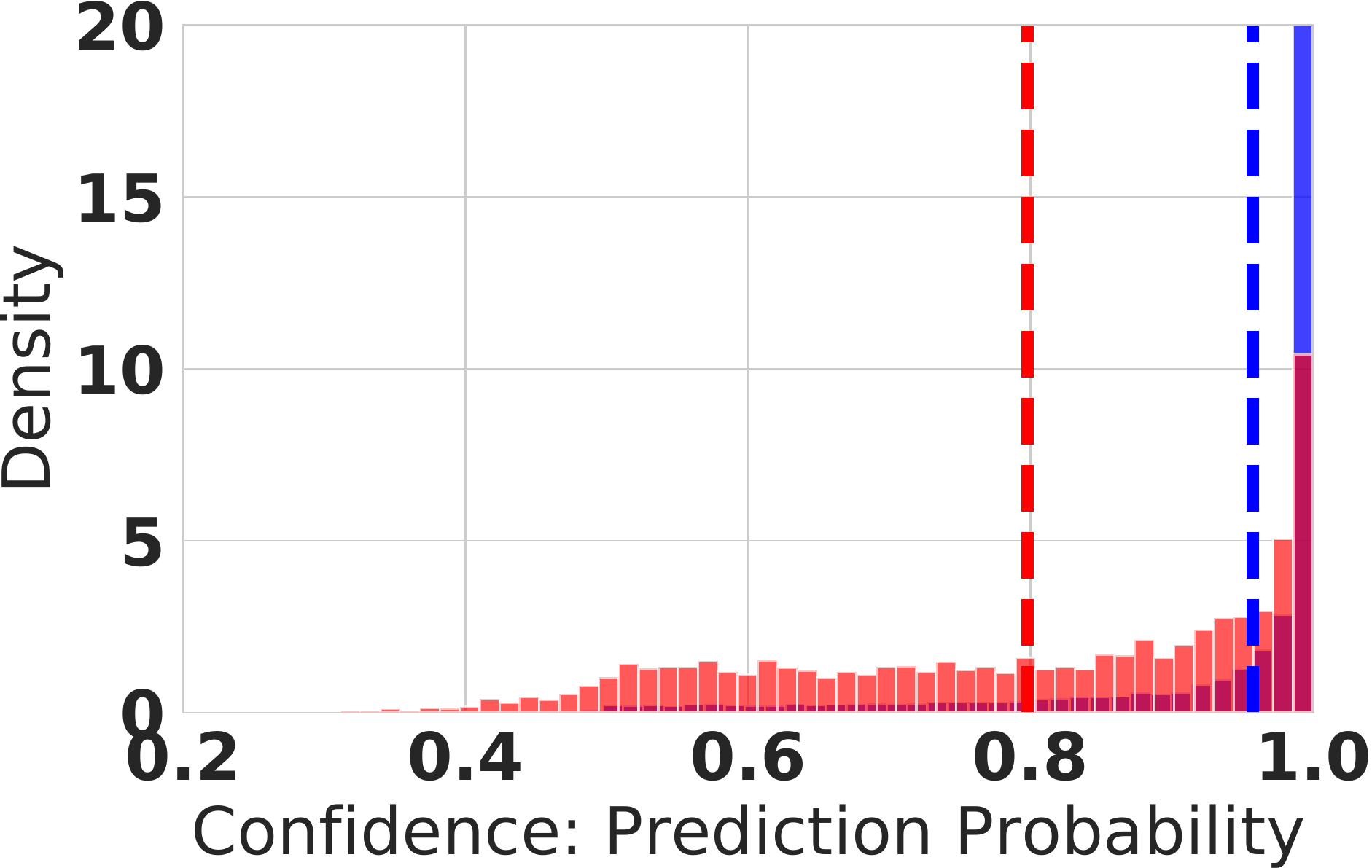}
  \caption{Baseline Env1-Test}
\end{subfigure}
\hfill
\begin{subfigure}{.492\columnwidth}
  \centering
  \includegraphics[width=0.95\textwidth, clip]{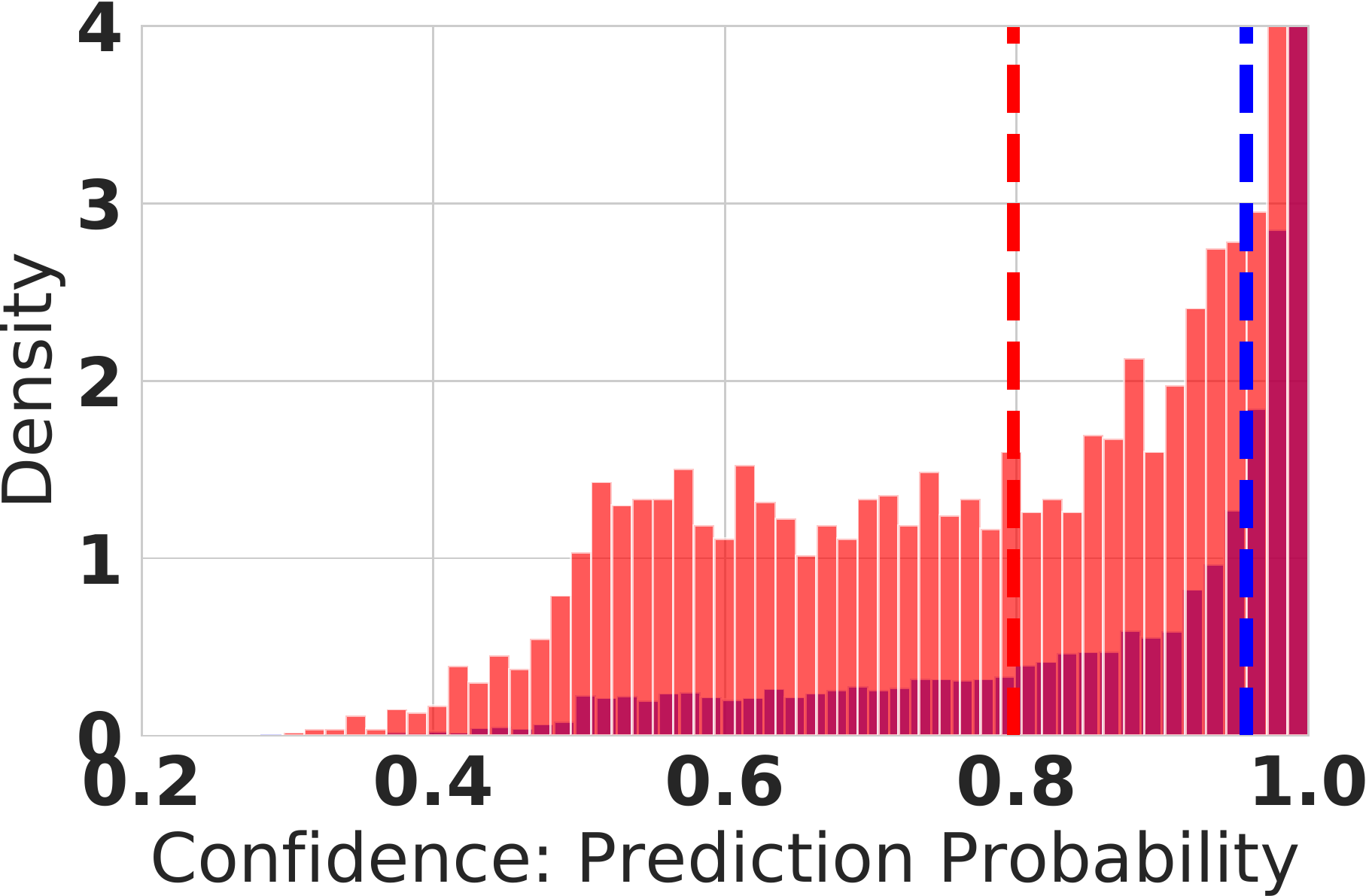}
  \caption{Baseline Env1-Test (Zoomed)}
\end{subfigure}
\\
\begin{subfigure}{.492\columnwidth}
  \centering
  \includegraphics[width=0.95\textwidth, clip]{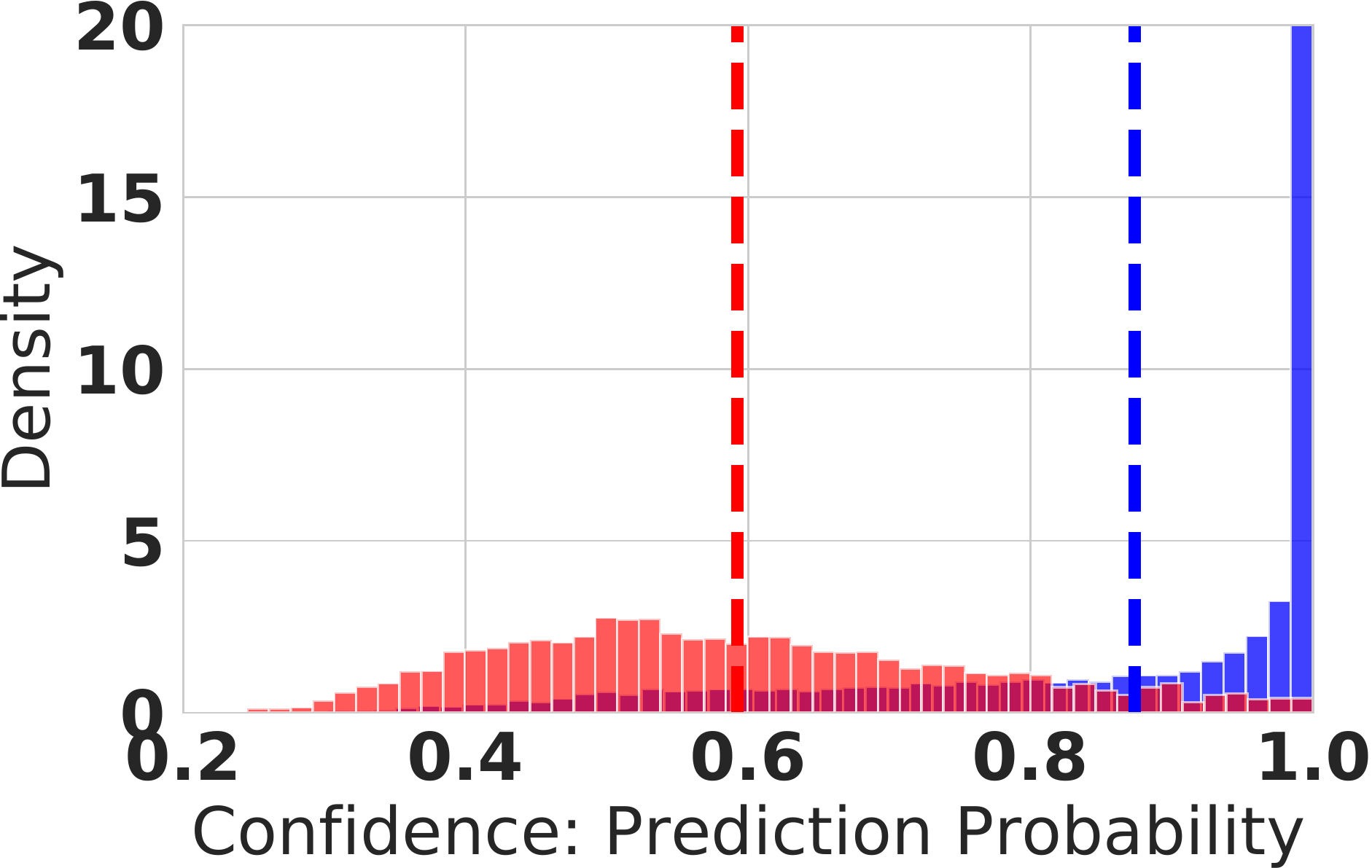}
  \caption{GP Env1-Test}
  \label{fig4b_cifar100_ace}
\end{subfigure}
\hfill
\begin{subfigure}{.492\columnwidth}
  \centering
  \includegraphics[width=0.95\textwidth, clip]{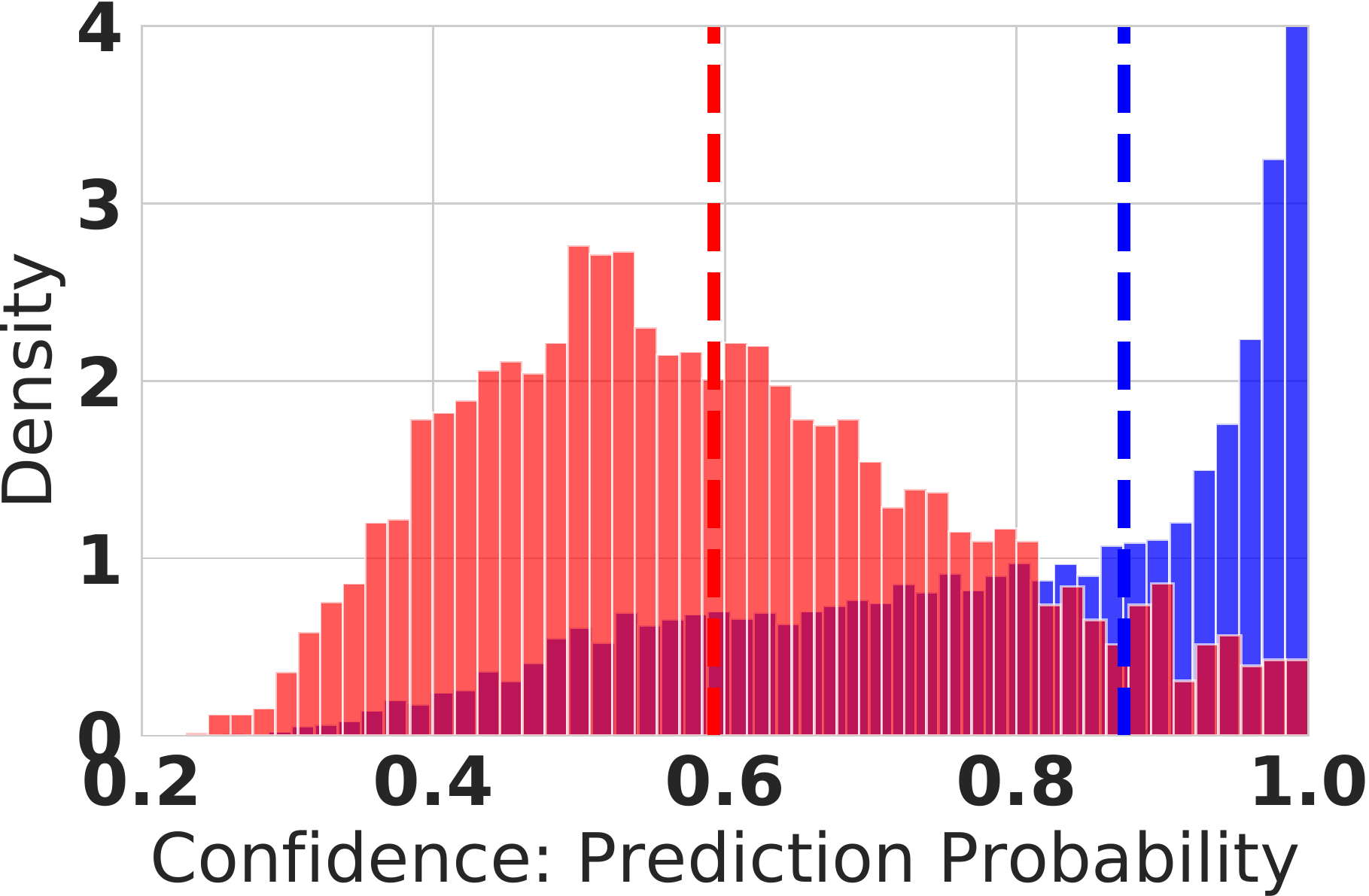}
  \caption{GP Env1-Test (Zoomed)}
\end{subfigure}
\caption{Confidence distribution of the top predicted class of Env1-Test by the Baseline network (a), and after GP scaling (c).  Panels (b) and (d) are zoom-ins of panels (a) and (c), respectively.  
The distribution of the correctly classified samples are depicted in blue and the mis-classified samples in red.
Even though the network correctly assigns high confidence for the correct samples (skewness of the blue distribution to the right), it also assigns high confidences for the mis-classified samples. 
GP uncertainty calibration remedies this by significantly lowering the confidence on mis-classified samples in the test set.  
The vertical dashed lines show the MMC of the respective distributions.
}
\label{fig:top1_confidences}
    \vspace{-0.50cm}
\end{figure}

\subsection{Over-confidence of Network Predictions}
While systems deployed for real-world tasks always require highly accurate predictions, we show the importance of detecting when these predictions are over-confident on wrong, unseen and unknown data.
In Fig.~\ref{fig:top1_confidences}, we depict the distribution of the confidences of the top predicted class, categorized into correct (blue) and incorrect (red) classifications.
Ideally, correctly classified samples should exhibit higher confidences than the incorrectly classified samples.
In Fig.~\ref{fig:top1_confidences}a, it can be seen that for the Baseline classifier most of the correctly classified samples have high confidences (skewness towards a confidence of $1.0$), but similar trends can also be seen for the incorrect samples, i.e. misclassifications are assigned relatively high confidences.  
This phenomenon is harmful as the model outputs confident predictions, regardless of whether the network was able to correctly recognize the object or not.

Using a post-hoc uncertainty calibration method such as GP~\cite{wenger2020calibration}, the severity of this harmful effect can be largely reduced (i.e. the red distribution is shifted significantly left). 

The dashed vertical lines in the figure show the Mean Max. Confidence (MMC) of the respective distributions, and comparing these it becomes clear that GP significantly reduces the confidences of most negative samples.
Even though post-hoc uncertainty calibration does not mitigate the problem, it still provides the first steps in the direction, and significantly improves the separation between confident correct and over-confident incorrect predictions.
We also can observe that in addition to lowering the confidences of the negative samples, GP also marginally reduces the confidences of the correct samples (i.e. slight left shift in the blue line), but this is not unexpected.
Being correct does not necessarily correlate with $100\%$ confidence, despite the labels indicating only a binary presence or absence of the object. 

It is important to note, that the labels are mere approximations of the object class but do not reflect the true ambiguity (labels indicate presence of object regardless of the amount of noise present or difficulty in recognizing the objects).
Given that measurement noise can have different severities and DL classifers could still recognize the objects with lower SNR, the confidence for such correct predictions should still be lower than $100\%$ as there is higher ambiguity.

\subsection{Uncertainty Calibration}

\begin{figure}[!t]
\centering
	\begin{subfigure}[t]{0.49\columnwidth}
		\includegraphics[width=0.95\textwidth]{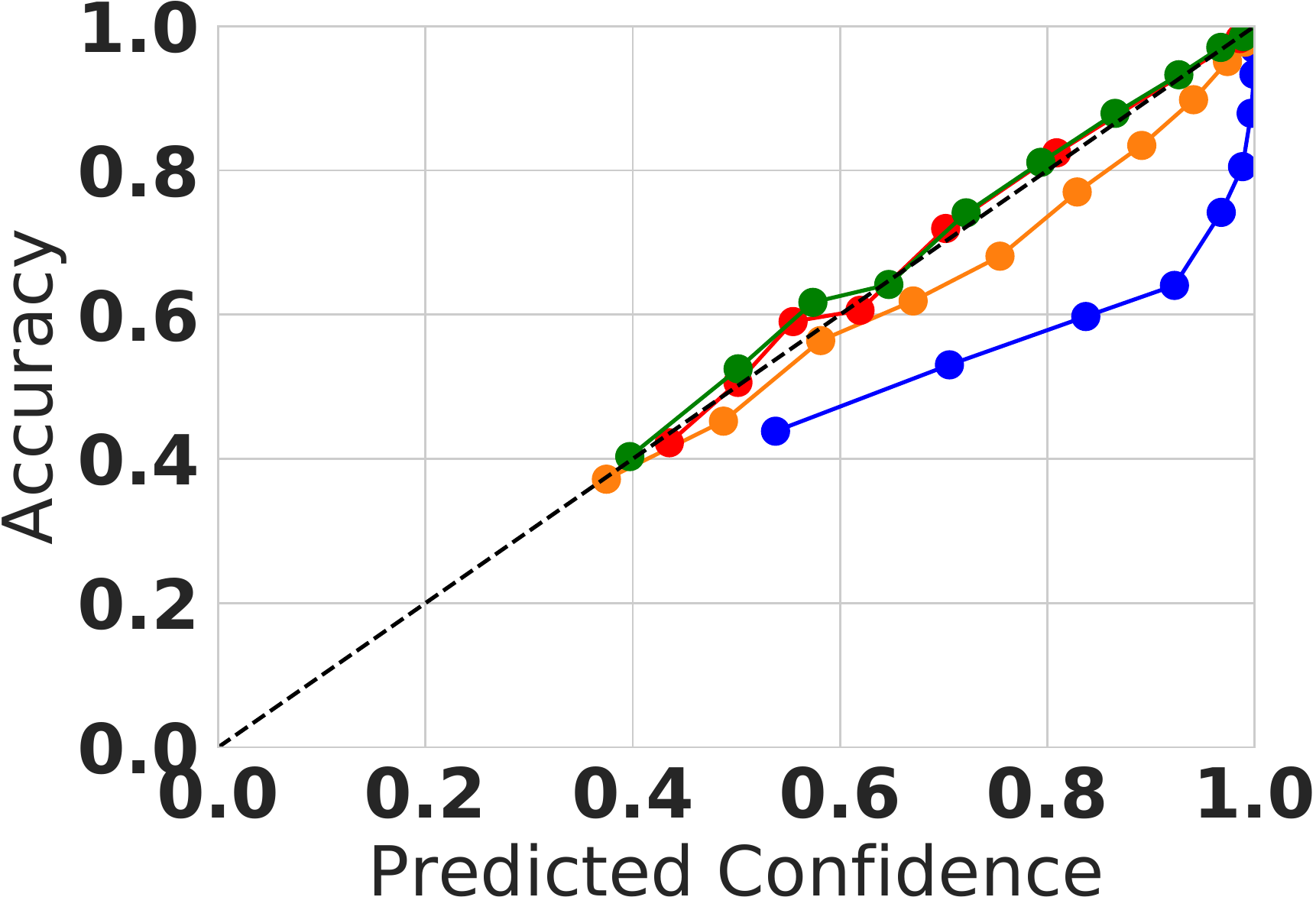}
		\caption{Env1-Test}
	\end{subfigure}
	\hfill
	\begin{subfigure}[t]{0.49\columnwidth}
		\includegraphics[width=0.95\textwidth]{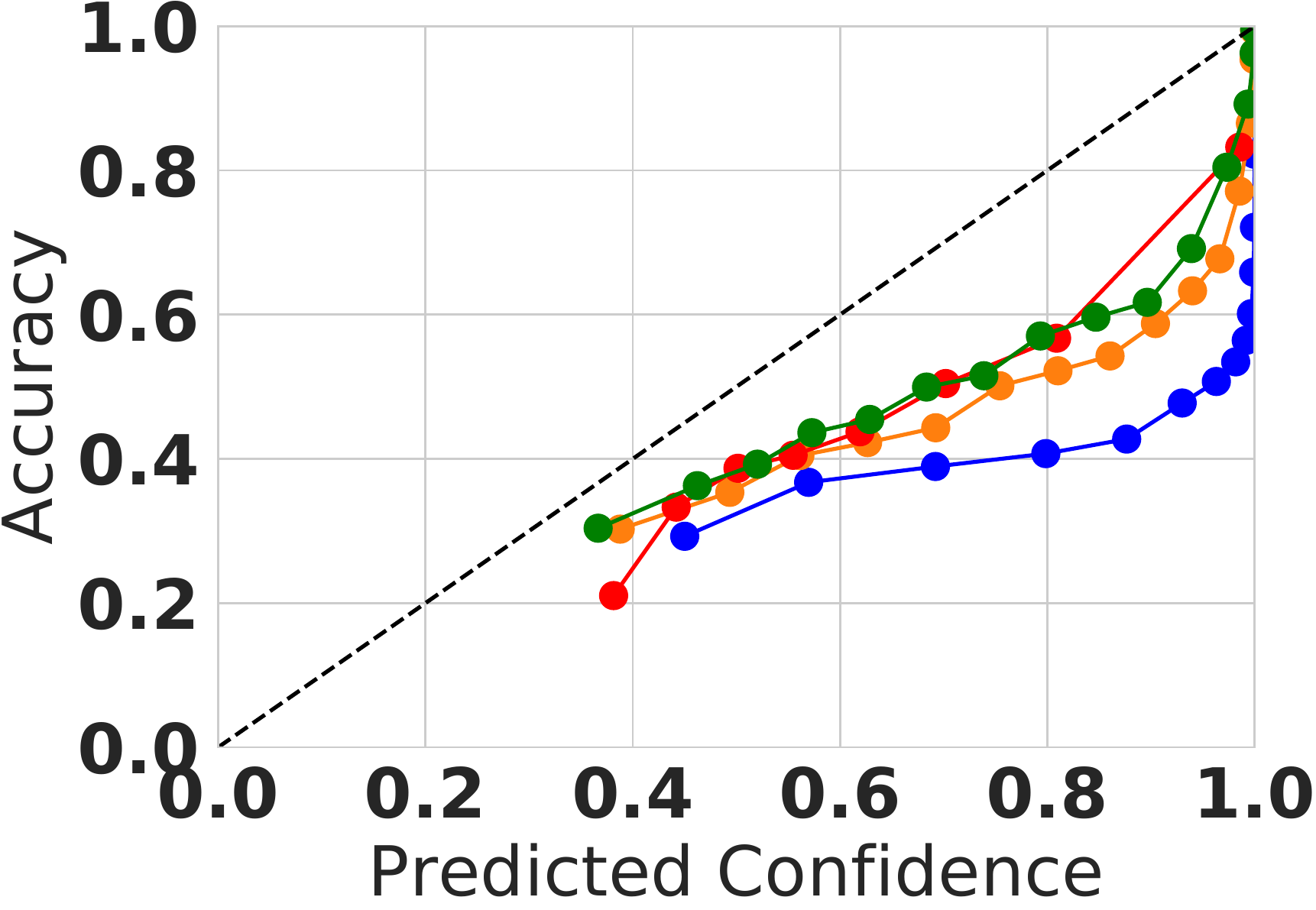}
		\caption{Env2-Test}
	\end{subfigure}
	\\
	\begin{subfigure}{0.75\columnwidth}
  \centering
  \includegraphics[width=1.0\textwidth, clip]{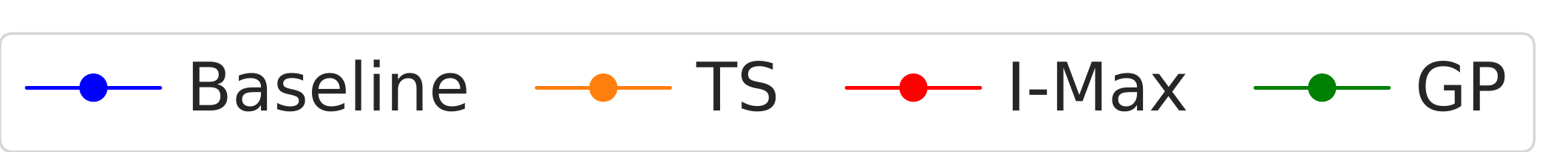}
  \label{legend}
\end{subfigure}
\caption{Reliability diagrams of the Baseline and uncertainty calibration methods. 
The baseline network confidences are severely over-confident for both test sets (larger distance to the diagonal line). 
The uncertainty calibration methods help to improve this by reducing the confidence.
Among these, GP and I-Max show the best calibration performance.}
\label{fig:cal_curves}
    \vspace{-0.90cm}
\end{figure}

In Fig.~\ref{fig:cal_curves}, we depict reliability diagrams for the two test datasets. 
These diagrams indicate how well the predicted confidences correspond to the true accuracy, i.e. how calibrated the uncertainty estimates are.
A perfectly calibrated network yields the black dashed $y=x$ line; curves below reflect over-confidence, and curves above under-confidence.
While the Baseline network outputs severely over-confident predictions, the post-hoc calibration methods significantly improve the calibration.  
Simple methods like TS help reduce the confidence, but further improvements can be made using more sophisticated methods such as GP and I-Max.
However even after uncertainty calibration, the predictions on Env2-Test are still overly confident. 
This shows that when encountered with realistic unseen changes, that the predictive accuracy and calibration become worse and uncalibrated predictions make it harder to detect such situations where the network is unsure.
The calibration performance is better, and significantly improved, for Env1-Test as it more similar to the validation/training set distributions than Env2-Test (i.e. a bias towards the original training set).
Calibration performance for Env2-Test, which is far more challenging, can improve with more training samples.

\begin{table*}
\normalsize
\centering
\caption{
Comparison between the Baseline and uncertainty calibration methods on Accuracy, ECE and MMC.
Despite Env2-Test being significantly harder (with a difference scene, object instances and driving patterns), the Baseline networks still show highly-accurate prediction ability.
We observe that all post-hoc methods help improve on the uncertainty metrics compared to the baseline, whereas accuracy remains similar.
GP offers the best over-all performance (with similar performance by I-Max), though it involves multiple stochastic forward passes and is not feasible on a real-time system.
Alternatively, I-Max offers competitive uncertainty calibration improvements and is significantly faster; allowing easy integration into any real-time system.
}
\label{tab:main_table}
\begin{tabular}{l|c|ccc|ccc}
\toprule
\rowcolor{lightgray}
				   &    & \multicolumn{3}{c|}{Env1-Test} &       \multicolumn{3}{c}{Env2-Test} \\
\rowcolor{lightgray}				        
      Method &  Time ($\mu$s) $\downarrow$ &  Acc $\uparrow$ &  ECE $\downarrow$ &   $\text{MMC}_{\text{incorrect}}$ $\downarrow$ &   Acc $\uparrow$  & ECE $\downarrow$ & $\text{MMC}_{\text{incorrect}}$ $\downarrow$ \\
\midrule

    Baseline &  / &   \textbf{81.67} $\pm$ 0.22 &  0.117 $\pm$ 0.04 &  0.82 $\pm$ 0.08   &   59.40 $\pm$ 1.03 &  0.307 $\pm$ 0.04 &   0.85 $\pm$ 0.07 \\
          TS & 0.75 &   \textbf{81.67} $\pm$ 0.22 &   0.064 $\pm$ 0.04 &   0.70 $\pm$ 0.08 &      59.40 $\pm$ 1.03 &  0.224 $\pm$ 0.04 &   0.74 $\pm$ 0.06 \\
 I-Max &  0.76   &  81.60 $\pm$ 0.21 &  \textbf{0.011} $\pm$ 0.00 &  0.61 $\pm$ 0.01 &    59.15 $\pm$ 1.16 &  0.155 $\pm$ 0.01 &  0.66 $\pm$ 0.01  \\
          GP & 35.3  &  \textbf{81.67} $\pm$ 0.22 &  0.019 $\pm$ 0.00 &   \textbf{0.59} $\pm$ 0.01 &    \textbf{59.41} $\pm$ 1.03 &  \textbf{0.134} $\pm$ 0.02 &  \textbf{0.62} $\pm$ 0.02\\
\bottomrule
\end{tabular}
    \vspace{-0.40cm}
\end{table*}

In order to quantitatively show the benefit of uncertainty calibration methods, we examine the accuracy, ECE, and MMC.  
These results, evaluated on the two test datasets (Env1-Test and Env2-Test) are shown in Tab.~\ref{tab:main_table}. 

Firstly, it is noted that on all metrics, significantly better performance is achieved on Env1-Test than on Env2-Test. 
This is owed to the fact that the classifier can easily generalize to Env1-Test, as it shares significant similarities to the training set Env1-Train.
However, it is important to note the impressive learning capabilities of the Baseline classifier to recognize most of the observations from Env2-Test.
These include measurements from a novel scene, some novel object instances (e.g. unseen car model) and viewing angles.
We observe that all uncertainty calibrations methods help improve on the uncertainty metrics (ECE, MMC) compared to the Baseline.
Even a simple scaling (e.g. TS) of all confidences (correct and incorrect) shows significant improvements.
Larger gains are observed for GP and I-Max, with the former offering slightly better performance.
As GP involves multiple stochastic forward passes (preventing its use in real-time systems), I-Max has large advantages in terms of test time, allowing easy integration into real-time systems with negligible extra computation time relative to the DL classifier.

\subsection{Spectra Corruption}
\begin{figure*}[!t]
\begin{minipage}{1.0\textwidth}
    \resizebox{1.0\textwidth}{!}{
\begin{tikzpicture}[scale=1.0\textwidth]
    \node (corruptions)[draw=white, line width=0.5mm, inner sep=0]{\includegraphics[width=2.0\textwidth]{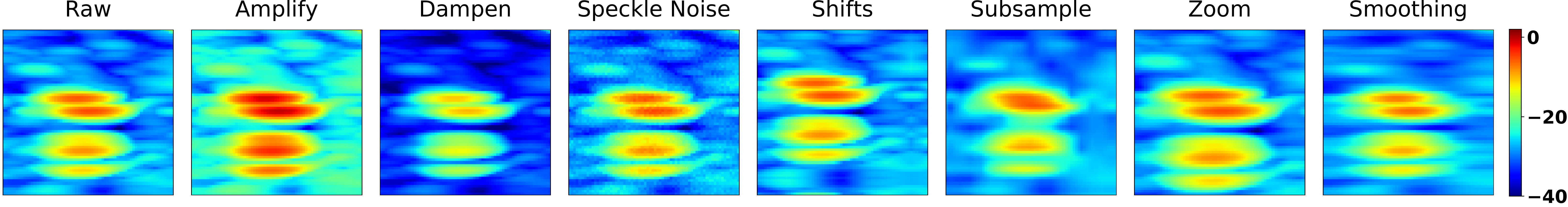}};
    \node(rlabel)[left=of corruptions, node distance=0cm, rotate=90, anchor=center,yshift=-0.5cm,font=\color{black}, scale=2] {\large Range [m]};
    \node(azlabel)[below =of corruptions, node distance=0cm, rotate=0, anchor=center,yshift=+0.6cm,font=\color{black}, scale=2] {\large Azimuth [${}^{\circ}$]};
 \end{tikzpicture}
    }
\end{minipage}
    \caption{Construction Barrier ROI spectra (in dB) after corrupting the Raw signal (left) with various corruptions. 
    }
\label{fig:corruption_samples}
    \vspace{-0.50cm}
\end{figure*}

Even though Env2-Test serves as a test for the classifier's ability to generalize to realistic changes in the spectra (due to the environment conditions changing), it does not provide a framework for incrementally changing the spectra in a controlled setting.
In order to study the behavior of the classifier to slight changes in the spectra, we incrementally corrupt the ROI spectra to measure the robustness of the radar networks to perturbations in the input.
In Fig.~\ref{fig:corruption_samples}, we depict some examples of the corrupted spectra used.
The corruptions were specifically chosen to be as realistic as possible to feasibly be encountered in a real-world system.
Though, we note that despite these not accurately simulating real-world corruptions, they are still well suited for the task of studying the uncertainty to unseen changes in the input of the classifier.

\begin{figure}[!t]
\centering
\begin{subfigure}{0.75\columnwidth}
  \includegraphics[width=1.0\textwidth, clip]{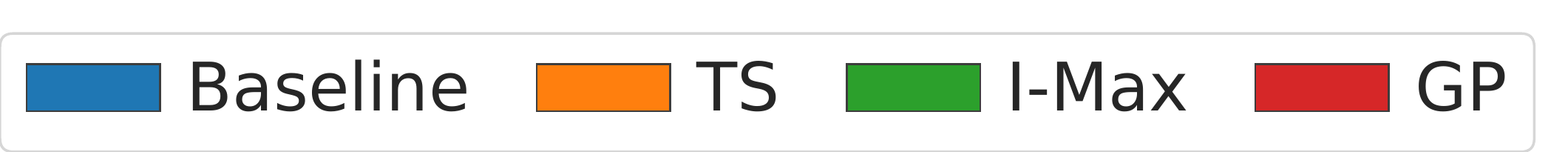}
  \label{legend}
\end{subfigure}
\begin{subfigure}[t]{0.492\columnwidth}
		\includegraphics[width=0.95\textwidth]{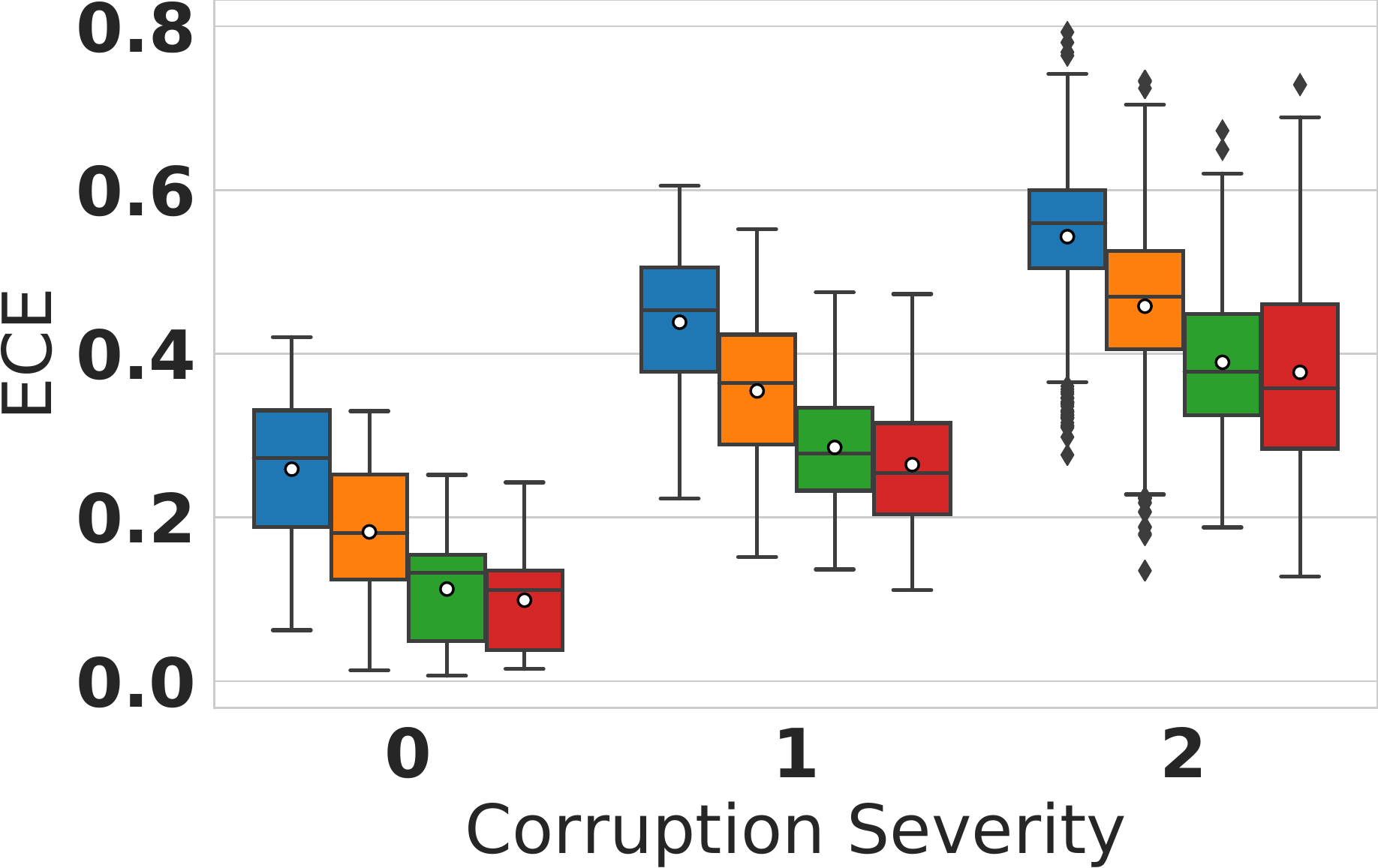}
		\caption{}
\end{subfigure}
\begin{subfigure}[t]{0.492\columnwidth}
		\includegraphics[width=0.95\textwidth]{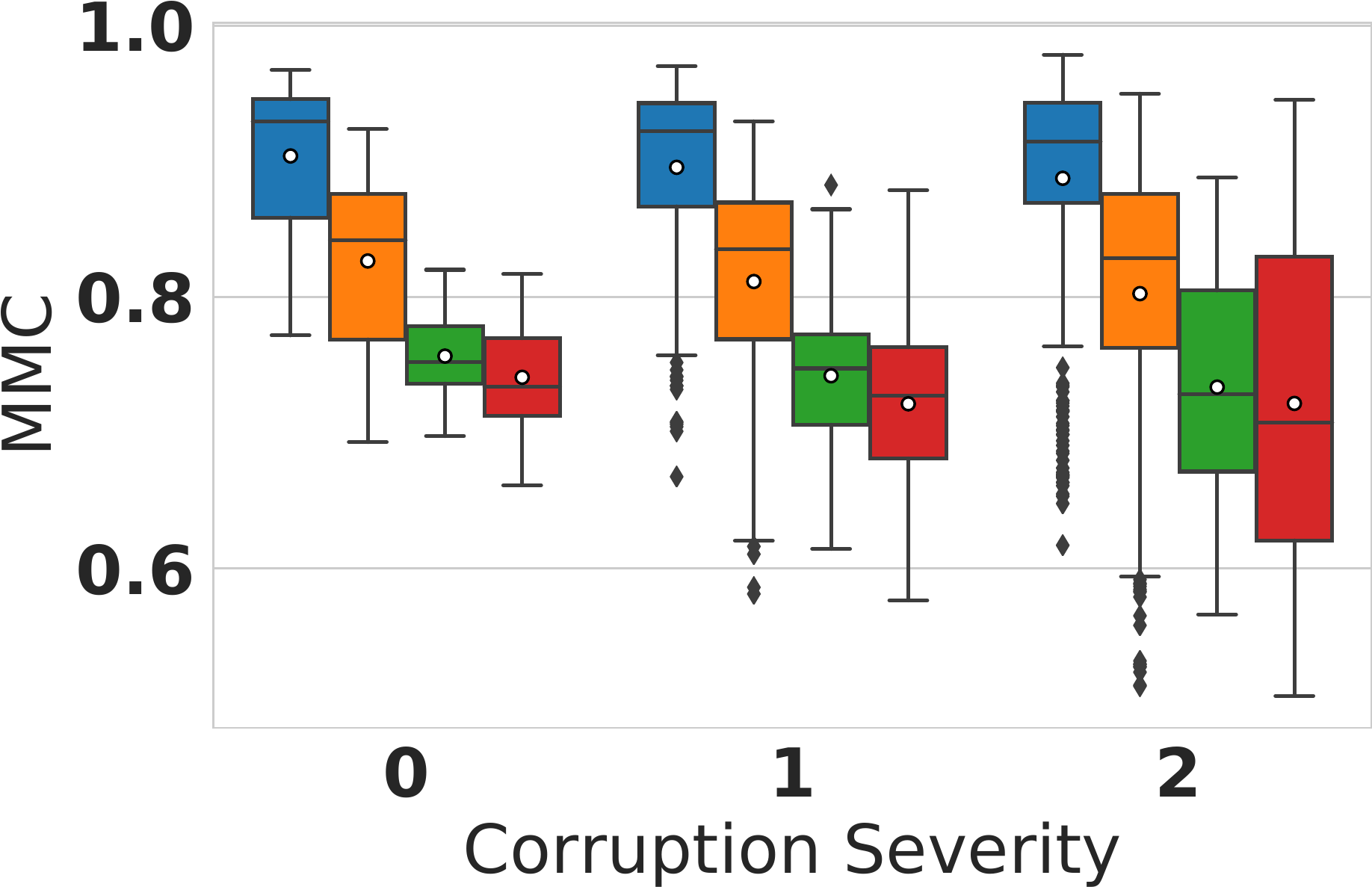}
		\caption{}
\end{subfigure}
\caption{Boxplot of Top 1 ECE errors of corrupted samples. All corruptions have been averaged out per severity. As expected, the Top 1 ECE becomes worse with higher severity and for each case all post-hoc methods improve the calibration performance. This means that the uncertainty estimates associated for these corruption samples are more accurate than the Baseline network confidences.}
\label{fig:boxplot_corruptions}
    \vspace{-0.50cm}
\end{figure}

All corruptions are tested at 3 severity levels: the baseline accuracy (averaged across all corruptions) of the networks at these three severities are 75.85\%, 51.31\%, and 37.33\%.
In Fig.~\ref{fig:boxplot_corruptions}a, we highlight the effect of the corruptions on the classifier's confidence estimates. 
As the corruptions get more severe, the ECE increases. 
A main factor to this increase in calibration error is illustrated in Fig.~\ref{fig:boxplot_corruptions}b, which shows the MMC over the same corruption severities.
The network fails to reduce its confidences despite becoming less accurate.
In summary, we again observe the trend of over-confidence, and, as before, the uncertainty calibration methods can yield significant improvements relative to the Baseline.

\subsection{Outlier Detection}

\begin{figure}[!h]
	\includegraphics[width=0.90\columnwidth]{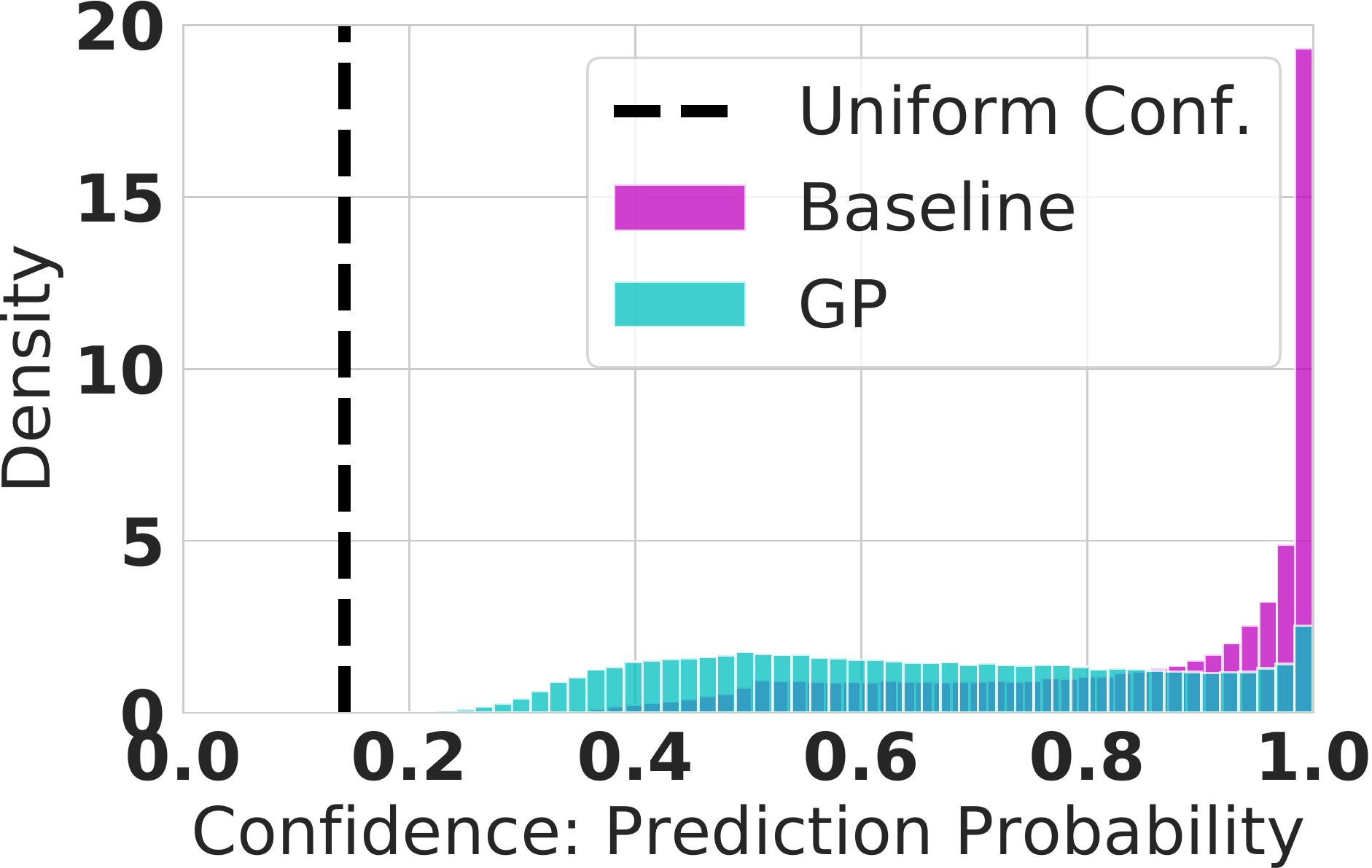}
	\caption{
	Even for the outlier objects the classifier exhibits extreme over-confidence.
	The desired confidence is achieved with a uniform confidence across all classes (dashed line).
}
	\label{fig:top1_confidences_OOD}
    \vspace{-0.55cm}
\end{figure}

To test even further the limits of the classifiers confidence to identify cases when it should be uncertain, we also evaluate on outlier data (i.e. objects never seen during training), for which it is impossible to classify correctly.
The ideal prediction in this case should be a uniform prediction over all classes, or low confidence to the most similar class, however, the network still assigns very high confidences (Fig.~\ref{fig:top1_confidences_OOD}).
As rare unseen object types can commonly be encountered by a real-world system, assigning lower confidences to them is vital; a classifier should not confidently predict on what it does not know. 
However, this is far from the case in the tested experimental setup.
In line with the previous results, post-hoc methods (GP) can significantly reduce these confidences.

\section{Conclusion}
As we observe rapidly growing interest in using deep learning for radar solutions, our intention is to study the behavior of radar networks beyond pure classification accuracy.
Ultimately neural networks are optimized for high accuracy, but our results show that it is important to also critically investigate the associated prediction confidences.
Through the use of measurements from an environment unfamiliar to the networks, we show that they fail to reflect higher uncertainties to unseen but realistic real-world changes in the spectra.
In an attempt to incrementally induce ambiguity to the spectra, a controlled set of corruptions presented to the networks, showed that confidences remained high, although predictions became highly inaccurate.
Lastly, we mimic realistic real-world encounters by attempting to classify unknown objects.
Clearly, no neural network can correctly classify these objects, but remarkably the confidence in the necessarily wrong predictions remained very high.
Displaying that highly-accurate networks fail to reflect uncertainty in their predictions, shows that more work is required for DL-based radar solutions to address the challenges of real-world automotive applications.
This should not diminish the fact that DL-based models are highly attractive because of their accuracy, and clearly learn important features, which are hard to model otherwise.
We conclude that future work should not only address increasing accuracy, but should guide  classifiers towards using their expressive powers for learning to model reliable uncertainty values.

As a first solution, we proposed using uncertainty calibration methods, and showed how these can effectively improve confidence calibration across multiple setups, and reduce the severe over-confidence to some extent.
Next steps could address the simulation of even more realistic corruptions and augmentations to further improve the evaluation and support the development of highly-accurate and calibrated solutions.

\bibliographystyle{IEEEtran}
\bibliography{literature}

\begin{thebibliography}{10}
\providecommand{\url}[1]{#1}
\csname url@samestyle\endcsname
\providecommand{\newblock}{\relax}
\providecommand{\bibinfo}[2]{#2}
\providecommand{\BIBentrySTDinterwordspacing}{\spaceskip=0pt\relax}
\providecommand{\BIBentryALTinterwordstretchfactor}{4}
\providecommand{\BIBentryALTinterwordspacing}{\spaceskip=\fontdimen2\font plus
\BIBentryALTinterwordstretchfactor\fontdimen3\font minus
  \fontdimen4\font\relax}
\providecommand{\BIBforeignlanguage}[2]{{%
\expandafter\ifx\csname l@#1\endcsname\relax
\typeout{** WARNING: IEEEtran.bst: No hyphenation pattern has been}%
\typeout{** loaded for the language `#1'. Using the pattern for}%
\typeout{** the default language instead.}%
\else
\language=\csname l@#1\endcsname
\fi
#2}}
\providecommand{\BIBdecl}{\relax}
\BIBdecl

\bibitem{patel_radar19}
K.~Patel, K.~Rambach, T.~Visentin, D.~Rusev, M.~Pfeiffer, and B.~Yang, ``Deep
  learning-based object classification on automotive radar spectra,'' in
  \emph{IEEE Radar Conference}, 2019.

\bibitem{pmlr-v70-guo17a}
C.~Guo, G.~Pleiss, Y.~Sun, and K.~Q. Weinberger, ``On calibration of modern
  neural networks,'' in \emph{ICML}, Sydney, Australia, Aug. 2017, pp.
  1321--1330.

\bibitem{hein2019relu}
M.~Hein, M.~Andriushchenko, and J.~Bitterwolf, ``Why {ReLU} networks yield
  high-confidence predictions far away from the training data and how to
  mitigate the problem,'' in \emph{Proceedings of the IEEE Conference on
  Computer Vision and Pattern Recognition}, 2019, pp. 41--50.

\bibitem{kristiadi2020being}
A.~Kristiadi, M.~Hein, and P.~Hennig, ``Being {Bayesian}, even just a bit,
  fixes overconfidence in {ReLU} networks,'' \emph{arXiv preprint
  arXiv:2002.10118}, 2020.

\bibitem{rohlingCFAR}
H.~Rohling, ``Ordered statistic cfar technique - an overview,'' in
  \emph{International Radar Symposium (IRS)}, Sept 2011, pp. 631--638.

\bibitem{schubert2015-dbscan}
E.~{Schubert}, F.~{Meinl}, M.~{Kunert}, and W.~{Menzel}, ``Clustering of high
  resolution automotive radar detections and subsequent feature extraction for
  classification of road users,'' in \emph{International Radar Symposium
  (IRS)}, 2015, pp. 174--179.

\bibitem{Scheiner_2019}
N.~Scheiner, N.~Appenrodt, J.~Dickmann, and B.~Sick, ``Radar-based road user
  classification and novelty detection with recurrent neural network
  ensembles,'' \emph{2019 IEEE Intelligent Vehicles Symposium (IV)}, Jun 2019.

\bibitem{prophet2018_car_dog_ped_classifier}
R.~{Prophet}, M.~{Hoffmann}, M.~{Vossiek}, C.~{Sturm}, A.~{Ossowska},
  W.~{Malik}, and U.~{L\"ubbert}, ``Pedestrian classification with a 79 {GHz}
  automotive radar sensor,'' in \emph{International Radar Symposium (IRS)},
  2018, pp. 1--6.

\bibitem{kraus2020GhostImagesRadar}
F.~Kraus, N.~Scheiner, W.~Ritter, and K.~Dietmayer, ``Using machine learning to
  detect ghost images in automotive radar,'' 2020.

\bibitem{schumann2018SemanticSegmentationRadar}
O.~Schumann, M.~Hahn, J.~Dickmann, and C.~Wohler, ``Semantic segmentation on
  radar point clouds,'' in \emph{International Conference on Information
  Fusion}, 07 2018, pp. 2179--2186.

\bibitem{ensembleLombacher2017}
J.~{Lombacher}, M.~{Hahn}, J.~{Dickmann}, and C.~{W\"ohler}, ``Object
  classification in radar using ensemble methods,'' in \emph{International
  Conference on Microwaves for Intelligent Mobility (ICMIM)}, March 2017, pp.
  87--90.

\bibitem{staticObjectClassificationDL-lombacher2016}
------, ``Potential of radar for static object classification using deep
  learning methods,'' in \emph{International Conference on Microwaves for
  Intelligent Mobility (ICMIM)}, 2016, pp. 1--4.

\bibitem{rohling-car-pedestrian-features-2010}
H.~{Rohling}, S.~{Heuel}, and H.~{Ritter}, ``Pedestrian detection procedure
  integrated into an 24 ghz automotive radar,'' in \emph{IEEE Radar
  Conference}, 2010, pp. 1229--1232.

\bibitem{bartsch2012-pedestrian-recognition}
A.~Bartsch, F.~Fitzek, and R.~Rasshofer, ``Pedestrian recognition using
  automotive radar sensors,'' \emph{Advances in Radio Science}, 2012.

\bibitem{prophet-image-based-pedestrian-classification2018}
R.~{Prophet}, M.~{Hoffmann}, A.~{Ossowska}, W.~{Malik}, C.~{Sturm}, and
  M.~{Vossiek}, ``Image-based pedestrian classification for 79 {GHz} automotive
  radar,'' in \emph{European Radar Conference (EuRAD)}, 2018, pp. 75--78.

\bibitem{convLSTM-Radar-microDoppler2019}
H.-U.-R. Khalid, S.~Pollin, M.~Rykunov, A.~Bourdoux, and H.~Sahli,
  ``Convolutional long short-term memory networks for {Doppler}-radar based
  target classification,'' in \emph{IEEE Radar Conference}, 2019.

\bibitem{objectDetectionRadarSpectra2020}
X.~{Dong}, P.~{Wang}, P.~{Zhang}, and L.~{Liu}, ``Probabilistic oriented object
  detection in automotive radar,'' in \emph{Conference on Computer Vision and
  Pattern Recognition Workshops (CVPRW)}, June 2020, pp. 458--467.

\bibitem{3dRadarcube-Cnn-classification2020}
A.~{Palffy}, J.~{Dong}, J.~F.~P. {Kooij}, and D.~M. {Gavrila}, ``{CNN} based
  road user detection using the {3D} radar cube,'' \emph{IEEE Robotics and
  Automation Letters}, vol.~5, no.~2, pp. 1263--1270, April 2020.

\bibitem{DropoutGalG16}
Y.~Gal and Z.~Ghahramani, ``Dropout as a {Bayesian} approximation: Representing
  model uncertainty in deep learning,'' in \emph{ICML}, 2016.

\bibitem{EnsemblesNIPS2017}
B.~Lakshminarayanan, A.~Pritzel, and C.~Blundell, ``Simple and scalable
  predictive uncertainty estimation using deep ensembles,'' in \emph{NeurIPS},
  2017.

\bibitem{Postels_2019_ICCV}
J.~Postels, F.~Ferroni, H.~Coskun, N.~Navab, and F.~Tombari, ``Sampling-free
  epistemic uncertainty estimation using approximated variance propagation,''
  in \emph{ICCV}, 2019.

\bibitem{patel2019onmanifold}
K.~Patel, W.~Beluch, D.~Zhang, M.~Pfeiffer, and B.~Yang, ``On-manifold
  adversarial data augmentation improves uncertainty calibration,'' in
  \emph{2020 26th International Conference on Pattern Recognition
  (ICPR)}.\hskip 1em plus 0.5em minus 0.4em\relax IEEE Computer Society, 2020.

\bibitem{OnMixupTrainThul}
S.~Thulasidasan, G.~Chennupati, J.~Bilmes, T.~Bhattacharya, and S.~Michalak,
  ``On mixup training: Improved calibration and predictive uncertainty for deep
  neural networks,'' \emph{NIPS}, 2019.

\bibitem{wenger2020calibration}
J.~Wenger, H.~Kjellstr{\"o}m, and R.~Triebel, ``Non-parametric calibration for
  classification,'' in \emph{AISTATS}, 2020.

\bibitem{patel2020imax}
K.~Patel, W.~Beluch, B.~Yang, M.~Pfeiffer, and D.~Zhang, ``Multi-class
  uncertainty calibration via mutual information maximization-based binning,''
  in \emph{International Conference on Learning Representations (ICLR)}, 2021.

\bibitem{lecun1995convolutional}
Y.~Lecun and Y.~Bengio, ``Convolutional networks for images, speech, and
  time-series,'' in \emph{The Handbook of Brain Theory and Neural
  Networks}.\hskip 1em plus 0.5em minus 0.4em\relax MIT Press, 1995.

\bibitem{batchnorm}
S.~Ioffe and C.~Szegedy, ``Batch normalization: Accelerating deep network
  training by reducing internal covariate shift,'' in \emph{International
  Conference on Machine Learning}, 2015, pp. 448--456.

\bibitem{dropout}
N.~Srivastava, G.~Hinton, A.~Krizhevsky, I.~Sutskever, and R.~Salakhutdinov,
  ``Dropout: A simple way to prevent neural networks from overfitting,''
  \emph{Journal of Machine Learning Research}, vol.~15, pp. 1929--1958, 2014.

\end{thebibliography}

\end{document}